\pdfoutput=1

\documentclass[11pt]{article}

\usepackage{naacl2021}

\usepackage{times}
\usepackage{latexsym}

\usepackage[T1]{fontenc}

\usepackage[utf8]{inputenc}

\usepackage{microtype}
\usepackage{url}
\usepackage{graphicx}
\usepackage{amsmath}
\usepackage{algorithm}
\usepackage{algpseudocode}
\usepackage{mathrsfs}
\usepackage{multirow}
\title{Lifelong Learning of Hate Speech Classification on Social Media}

\author{Jing Qian$^\dagger$, 
Hong Wang$^\dagger$, 
Mai ElSherief$^\ast$, 
Xifeng Yan$^\dagger$\\
$^\dagger$ University of California, Santa Barbara\\
$^\ast$ Georgia Institute of Technology \\
  {\tt \{jing\_qian,hongwang600,xyan\}@cs.ucsb.edu} \\ \tt mai.h.sherief@gmail.com  \\}
\begin{document}
\maketitle
\begin{abstract}
Existing work on automated hate speech classification assumes that the dataset is fixed and the classes are pre-defined. However, the amount of data in social media increases every day, and the hot topics changes rapidly, requiring the classifiers to be able to continuously adapt to new data without forgetting the previously learned knowledge. This ability, referred to as lifelong learning, is crucial for the real-word application of hate speech classifiers in social media. In this work, we propose lifelong learning 
of hate speech classification on social media. To alleviate catastrophic forgetting, we propose to use Variational Representation Learning (VRL) along with a memory module based on LB-SOINN (Load-Balancing Self-Organizing Incremental Neural Network). Experimentally, we show that combining variational representation learning and the LB-SOINN memory module achieves better performance than the commonly-used lifelong learning techniques.
\end{abstract}
\section{Introduction}
\label{sec:intro}
With the rapid rise in user-generated web content, the scale and complexity of online hate have reached unprecedented levels in recent years.
ADL (Anti-Defamation League) conducted a nationally representative survey of Americans in December 2018 and the report shows that 
over half (53\%) of Americans experienced some type of online harassment.\footnote{https://www.adl.org/onlineharassment} This number is higher than the 41\% reported to a comparable question asked in 2017 by the Pew Research Center~\cite{pew2017online}. To address the growing online hate,
a great deal of research has focused on automatic hate speech classification.
 Most of the previous work focuses on binary classification~\cite{warner2012detecting, zhong2016content, nobata2016abusive, gao2017recognizing, qian2018leveraging} or coarse-grained multi-class classification~\cite{waseem2016hateful, badjatiya2017deep, davidson2017automated}.
 \begin{figure}[t]
    \centering
    \includegraphics[width=0.49\textwidth]{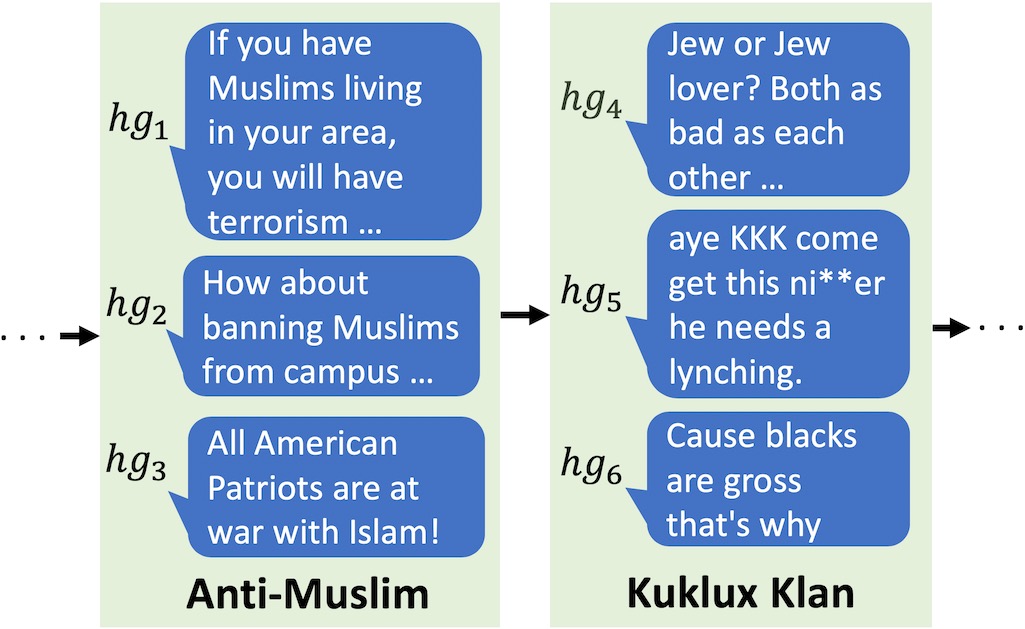}
    \caption{An illustration of our proposed task. $hg_i$: the $i$th hate group. The model is trained on a sequence of sub-datasets, split by their hate ideologies, e.g., anti-Muslim and Kuklux Klan. The task on each sub-dataset is to identify the hate group given the tweet.}
    \label{fig:intro}
\end{figure}
\citet{qian2018hierarchical} argue that fine-grained classification is necessary for fine-grained hate speech analysis. The Southern Poverty Law Center (SPLC) monitors hate groups throughout the United States by a variety of methodologies to determine the activities of groups and individuals, including reviewing hate group publications.\footnote{https://www.splcenter.org/fighting-hate/extremist-files/ideology} Therefore, instead of differentiating normal posts from the other offensive ones, ~\citet{qian2018hierarchical} propose a more fine-grained hate speech classification task that attributes hate groups to individual tweets. 
However, a common limitation of all the research mentioned above is that they assume the dataset to be static and train the classifiers on each isolated dataset, i.e., isolate learning, ignoring the rapid increase of the amount of data in social media and the rapid change of the hot topic.

A report from L1ght\footnote{https://l1ght.com/Toxicity\_during\_coronavirus\_Report-L1ght.pdf}, a company that specializes in measuring online toxicity, suggests that amid the growing threat of the coronavirus, there has been a 900\% growth in hate speech towards China and Chinese people on Twitter since February 2020.
As a result of the rapid change of social media content, the hate speech classifiers are required to be able to continuously learn and accumulate  knowledge from a stream of data, i.e., lifelong learning. Learning on each portion of the data is considered as a task, so a stream of tasks are joined to be trained sequentially. In this work, we propose a novel lifelong fine-grained hate speech classification task, as illustrated in Figure~\ref{fig:intro}. 
The models trained by isolate learning tend to face catastrophic forgetting~\cite{mccloskey1989catastrophic,ratcliff1990connectionist,mcclelland1995there,french1999catastrophic} due to a non-stationary data distribution in lifelong learning. To address this problem, an extensive body of work has been proposed for various lifelong learning tasks.  However, our experiments show that the commonly-used lifelong learning methods still exhibit catastrophic forgetting in our proposed tasks. One important difference between the Twitter hate group dataset and the other image datasets commonly used in lifelong learning study is that the similarity among the different tasks is unstable and relatively low, as indicated by the low average Jaccard Indexes of the topic words in Table~\ref{tab:data}. To alleviate this problem, we introduce VRL to distill the knowledge from each task into a latent variable distribution.
We also augment the model with a memory module and adapt the clustering algorithm, LB-SOINN, to select the most important samples from the training dataset of each task. Our implementation is publicly available\footnote{https://www.aclweb.org/anthology/2021.naacl-main.183/}.

Our contributions are three-fold:
\begin{itemize}
    \item This is the first paper on lifelong learning of fine-grained hate speech classification.
    \item We propose a novel method that utilizes VRL along with an LB-SOINN memory module to alleviate catastrophic forgetting resulted from a severe change of data distribution.
    \item Experimental results show that our proposed method outperforms the state-of-the-art significantly on the average F1 scores.
\end{itemize}

\section{Related Work}
\label{sec:related}
Most research on lifelong learning alleviates catastrophic forgetting in the following three directions.

\noindent{\bf Regularization-based Methods:}
These methods impose constraints on the weight update. The goal of the constraints is to minimize deviation from trained weights when training on a new task. The constraints are generally modeled by additional regularization terms~\cite{ewc,Zenke2017,pathnet2017,liu2018rotate,ritter2018online}.
Elastic Weight Consolidation (EWC)~\cite{ewc} alleviates catastrophic forgetting by slowing down learning on the model parameters which are important to the previous task.  The importance of the parameters is estimated by the Fisher information matrix.  
Instead of the Fisher information matrix, PathNet~\cite{pathnet2017} uses agents embedded in the neural network to determine which parameters of the neural network can be reused for new tasks and the task-relevant pathways are frozen during training on new tasks.
\begin{table}[t!]
\small
\centering 
\begin{tabular}{|l|c|c|}
\hline
\textbf{Ideology} & \textbf{Avg. JI} & \textbf{Keywords} \\\hline
Christian Identity &0.019 &Jesus, Yahuwshua\\\hline
Radical Tr. Catholic &0.031 &catholic, remnant\\\hline
Neo Confederate &0.039 &southern, Free Dixie \\\hline
Anti Semitism &0.047 &Israel, Trump\\\hline
Anti Catholic &0.049 &Texe Marrs, truth\\\hline
Hate Music &0.049 &death, radio\\\hline
Anti Muslim  &0.064 &Muslim, Islam\\\hline
Black Separatist &0.071 &black, panther\\\hline
Racist Skinhead &0.074 &shirt, white\\\hline
Anti Immigration &0.075 &immigration, border\\\hline
Holocaust Identity &0.078 &Jewish, Trump\\\hline
Neo Nazi &0.091 &Hitler, white\\\hline
Kuklux Klan &0.100 &ni**a, f**king\\\hline
Anti LGBTQ &0.100 &family, marriage\\\hline
White Nationalist &0.105 &white, America\\\hline
\end{tabular}
\caption{Information about the 15 hate ideologies.  Tr.: Traditional. Avg JI: the average of the Jaccard Index between the topic words of one ideology and those of another ideology. The topic words are extracted by Latent Dirichlet Allocation (LDA)~\cite{blei2003latent}. The top 2 most frequent topic words are selected as keywords.}
\label{tab:data}
\end{table}

\noindent{\bf Architecture-based Methods:}
The main idea of this approach is to change architectural properties to dynamically accommodating new tasks, such as assigning a dedicated capacity inside a model for each task.
 ~\citet{progressive2016} propose Progressive Neural Networks, where the model architecture is expanded by allocating a new column of neural network for each new task.
 ~\citet{part2016incremental, part2017incremental} combine Convolutional Neural Network with LB-SOINN for incremental online learning of object classes. Although they also use LB-SOINN in their work, the usage of LB-SOINN in this work is completely different. They use LB-SOINN to predict object class while our proposed method adapts the original LB-SOINN to calculate the importance of the training samples without making any prediction on the class.
A problem with the methods in this category is that the available computational resources are limited in practice. As a result, the model expansion will be prohibited when the number of tasks increases to a certain degree. 

\noindent{\bf Data-based Methods:}
These methods alleviate catastrophic forgetting by utilizing a memory module, which either stores a small number of real samples from previous tasks or distills knowledge from previous tasks.
The main feature of Gradient Episodic Memory (GEM)~\cite{gem} is the episodic memory, storing a subset of the samples from the observed tasks.  GEM computes the losses on the episodic memories and treats them as inequality constraints, avoiding them to increase. Averaged GEM~\cite{agem} is a more efficient version of GEM. ~\citet{epidosicmemory} propose a lifelong language learning model using a key-value memory module for sparse experience replay and local adaptation. ~\citet{lamol} formulate lifelong language learning as a language modeling task and replay the generated pseudo-samples of previous tasks during training.

There are also studies combining multiples methods above.~\citet{xia2017distantly} 
combine the architecture-based method and the data-based method. ~\citet{wang2019sentence} combine the regularization method and the data-based method for lifelong learning on relation extraction. Our proposed method is also a combination of the regularization method and the data-based method but in a different way.

\section{Task Description}
\label{sec:desc}
We use the dataset as in~\citet{qian2018hierarchical}, where the tweet handles are collected based on the hate groups identified by SPLC. SPLC categorizes these hate groups according to their hate ideologies. For each hate ideology, the top three Twitter handles are selected in terms of the number of followers. 
The dataset includes all the content (tweets, retweets, and replies) posted with each Twitter account from the group's inception date, as early as 2009, until 2017. Altogether, the dataset consists of 42 hate groups from 15 different ideologies. Table~\ref{tab:data} shows the 15 ideologies. Each instance in the dataset is a text tuple of (tweet, hate group name, hate ideology).

We separate the dataset by ideology. The reason is that various existing hate speech datasets collect data using keywords or hashtags~\cite{waseem2016hateful,davidson2017automated,golbeck2017large}, which have a strong relationship with hate ideologies or topics. We also observe that the hot spots of society can lead to a significant shift of major hate speech topics or the emergence of new hate ideologies on social media as mentioned in section~\ref{sec:intro}, indicating that the expansion of the hate speech dataset may be accompanied by the emergence of new hate ideologies.

Therefore, we separate the collected data into a sequence of 15 subsets according to their ideologies
and sort them by the date of the first tweet post in each subset, from the earliest to the latest. 
The task on each subset is to identify the hate group given the tweet text. ~\citet{qian2018hierarchical} propose a hierarchical Conditional Variational Autoencoder model for the fine-grained hate speech classification task. The architecture and the training process of their model require the number of classes to be pre-defined. However, we do not pre-define the number of classes in our task since such kind of information is not available in the real-world application of lifelong learning. The model should be able to incorporate emerging hate groups at any time of training. In order to satisfy this condition, we formulate the task of identifying the group as a ranking task, instead of a classification task. For each tweet, we provide the model with a set of candidate groups, consisting of all the previously seen hate groups, including the ground truth group. The model takes each combination of the tweet and the candidate group as input and outputs a score. The corresponding loss function is:
\begin{equation}
\mathcal{L}_r\!=\!\!\sum_{(x,y_s)\in D}\sum_{y_{i}\in{Y\backslash\{y_{s}\}}}\!\!\!\!\!h(f_{\theta}(x, y_{s})-f_\theta(x, y_{i}))
\label{eq:lossmargin}
\end{equation}
where $x$ is the tweet text, $y_s$ is the ground truth group of $x$. $Y$ is candidate group set of $x$, which consists of all the seen hate groups until $x$ is observed by the model, including the ground truth group $y_s$ of $x$, so $y_i \in {Y\backslash\{y_{s}\}}$ is the negative candidate group of $x$. $f_{\theta}$ is the scoring model parameterized by $\theta$. $h(a)=\max(0, m-a)$, $m$ is the chosen margin.

Same as in other lifelong learning studies, we consider learning on each of the hate ideologies in the sequence as a task, so we have a sequence of 15 tasks.
As mentioned in section~\ref{sec:intro}, the similarity among our tasks is unstable and relatively low. Therefore, when the model is continuously trained on the tasks, it may encounter a sudden change of vocabulary, topic, and input data distribution. This makes our tasks more challenging compared to the other lifelong learning tasks because the abrupt change can make the catastrophic forgetting problem more severe. This is also the reason that some techniques achieving significant improvement in the image classification tasks do not perform well on our task (see section~\ref{sec:experiment}). 
\begin{figure*}[t]
\centering
\includegraphics[width=0.68\textwidth]{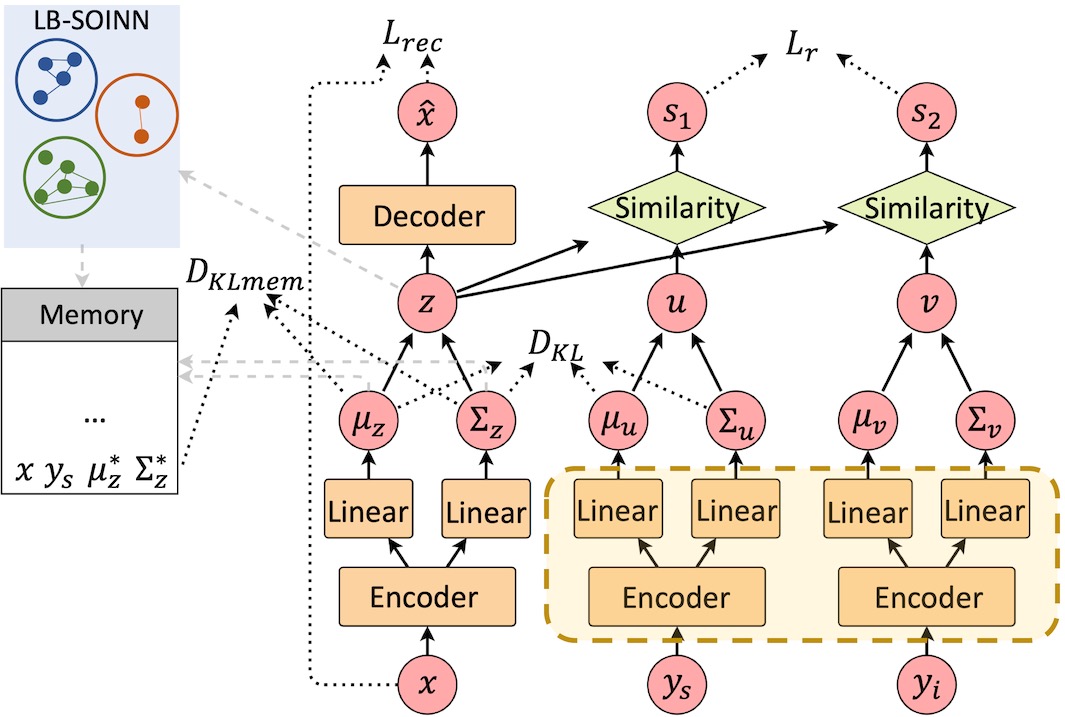}
\caption{An illustration of our method. The dotted arrows indicate the computation of the loss. The light-colored dashed arrows illustrate the update of the memory module. Note that the layers in the rounded rectangle share parameter weight. There is only one encoder for the group input, followed by two linear layers. We make a copy of it in the figure just for a clear illustration of loss computation. $\hat{x}$: the reconstructed tweet input. $s_1$, $s_2$: scores of ($x$, $y_s$) and ($x$, $y_i$) separately. $\mu_z^*$ and $\Sigma_z^*$ are the previously memorized distribution on the latent variable of $x$. $L_{rec}$ is the reconstruction loss, which is the first term in equation~\ref{equ:lossvae}. Please refer to section~\ref{sec:approach} for the meaning of other variables in the figure.}
\label{fig:model}
\end{figure*}
\section{Our Approach}
\label{sec:approach}
As mentioned in section~\ref{sec:related}, one way to alleviate catastrophic forgetting is to use a memory module, storing a small number of real samples from previous tasks and a simple way to utilize the memorized samples is to replay the memory when training on a new task, such as mixing them with the training samples from the current task. 
The idea behind this approach is that the memorized samples should reflect the data distribution so that the replay of the memory can help the model make invariant predictions on the samples of the previous tasks. However, this approach may not work well when the size of the memory is small. The reason is that when there is only a small amount of data memorized, the memory is not able to reflect the data distribution of the previous task and thus the model can easily overfit on the memorized samples instead of generalizing to all the samples in the previous task. 

We address this problem from two aspects. First, since the memory size is limited, it is beneficial to select the most representative training samples in the previous tasks to memorize. Second, simply storing the real training samples in the memory may not be sufficient to represent the knowledge of the previous tasks, so we need a better way to distill knowledge from the observed samples along with a method to utilize it when training on a new task. We combine two techniques: Variational Representation Learning (VRL) and Load-Balancing Self-Organizing Incremental Neural Network (LB-SOINN) to achieve these goals. We propose a supervised version of LB-SOINN to select the most important training samples in the current task. VRL not only distills the knowledge from the current training task but also provides an appropriate hidden representation as input for the LB-SOINN, so we introduce VRL first.

\subsection{Variational Representation Learning}
\label{subsec:vae}
The distilled knowledge of previous tasks can take various forms, but the key point is that it should be related to the data distribution of the corresponding task so that it can be utilized to alleviate catastrophic forgetting. Inspired by the Variational Autoencoder (VAE)~\cite{kingma2013auto}, we consider the distribution of the hidden representation of the input data as the distilled knowledge. 

The original VAE model is proposed for data generation, so the objective of the original VAE is:
\begin{equation}
     Obj=\sum_{x\in X}\log p(x)
\end{equation}
\begin{equation}
    p(x)=\int_{z}p(x|z)p(z)dz
\end{equation}
$z$ is the latent variable, i.e., the hidden representation of the input. Since the integration over $z$ is intractable, we instead try to maximize the corresponding evidence lower bound (ELBO) and the corresponding loss function is as follows:
\begin{equation}
\begin{aligned}
\mathcal{L}_{vae}=\sum_{x\in X}&E_{z\sim p_\alpha(z|x)}[-\log p_\varphi(x|z)]+\\
&D_{KL}[q_\alpha(z|x)||p_\beta(z)]
\label{equ:lossvae}
\end{aligned}
\end{equation}
$p(x|z)$, $q(z|x)$, and $p(z)$ are the likelihood distribution, posterior distribution, and prior distribution. $\alpha$,$\varphi$, and $\beta$ indicate parameterization. The loss function can be separated into two parts. The first part $E[-\log p(x|z)]$ is the reconstruction loss, trying to reconstruct the input text from the latent variable. It pushes $z$ to reserve as much information of the input as possible. This is consistent with our goal to learn the knowledge of the data distribution.
The second part is $D_{KL}[q(z|x)||p(z)]$, where $D_{KL}$ is the Kullback–Leibler (KL) divergence. Minimizing it pushes the posterior and the prior distributions to be close to each other. 
By assuming the posterior $p(z|x)$ to be a multivariate Gaussian distribution $\mathcal{N}(\mu_z, \Sigma_z)$, the latent variable $z$ is sampled from $\mathcal{N}(\mu_z, \Sigma_z)$. 

In the original VAE, $p(z)$ is chosen to be a simple Gaussian distribution $\mathcal{N}(0, 1)$. However, this is over-simplified in our task because different from the unsupervised generation task of the original VAE, our ranking task is supervised. Our task not only requires $z$ to contain information of the tweet text itself but also requires it to indicate the group information of the tweet. In other words, the distilled distribution should be conditioned on both the tweet and its group label to reflect the data distribution in a supervised task. Setting the prior to be the same for all the hate groups pushes $z$ or the distribution of $z$ to ignore the label information. Instead, the prior should be different for each hate group, so we replace $p(z)$ with $p(u|y_s)$, where $y_s$ is the group label of $x$ and $u$ is the latent variable. $p(u|y_s)$ is assumed to be a multivariate Gaussian distribution $\mathcal{N}(\mu_u, \Sigma_u)$. Note that the replacement itself can not guarantee $p(u|y_s)$ to be different for each hate group because the loss function in equation~\ref{equ:lossvae} does not push $p(u|y_s)$ to satisfy this condition. However, the ranking loss function~\ref{eq:lossmargin} fills in the gap. Therefore, our loss function on the current training task is a combination of these two.
\begin{equation}
\begin{aligned}
\mathcal{L}_{cur}\!=\!\!\!\!\!\!\sum_{(x,y_s)\in D} & \sum_{y_{i}\in{Y\backslash\{y_{s}\}}}\!\!\!\!\!h(f_{\theta}(x, y_{s})\!-\!f_\theta(x, y_{i}))\\
&\!\!\!+\!E_{z\sim p_\alpha(z|x)}[-\log p_\varphi(x|z)]\\
&\!\!\!+D_{KL}[q_\alpha(z|x)||p_\beta(u|y_{s})]
\end{aligned}
\end{equation}
The right part of Figure~\ref{fig:model} illustrates the computation process of VRL.

\subsection{LB-SOINN Memory Module}
\label{subsec:memory}
VRL provides a way to summarize knowledge into latent variable distributions. However, we still need a method to utilize the learned distribution to alleviate catastrophic forgetting. We do this by incorporating a memory module $D_{mem}$ to store a small subset of important training samples along with their latent variable distributions, so each sample stored in the memory is a tuple of $(x, y_z, q_{\alpha^\prime}(z|x))$. Here $q_{\alpha^\prime}(z|x)$ is the distribution computed when the model completes training on the task that $(x, y_z)$ belongs to. The memorized samples are taken as anchor points when training on a new task. We introduce a memory KL divergence loss to push $q_\alpha(z|x)$ computed when training on a new task to be close to the memorized distribution $q_{\alpha^\prime}(z|x))$. Therefore, the complete loss function is:
\begin{equation}
\begin{aligned}
&\mathcal{L}=\mathcal{L}_{cur}+ D_{KLmem}\\
&=\mathcal{L}_{cur}+\!\!\!\!\!\!\!\sum_{(x,y_s)\in D_{mem}}\!\!\!\!\!\!\!\!D_{KL}[q_\alpha(z|x)||q_{\alpha^\prime}(z|x))]
\end{aligned}
\label{eq:lossfinal}
\end{equation}

Since the size of the memory is limited, 
we introduce a supervised version of LB-SOINN to select the most important training samples in the current task. The input for the LB-SOINN is the hidden representation of the tweet text, which is $z$ in the case of Variational Representation Learning (see Figure~\ref{fig:model}). We refer readers to~\citet{zhang2013load} for the detailed explanation of LB-SOINN. The original LB-SOINN is an unsupervised clustering algorithm that clusters unlabeled data by topology learning. We utilize the topology learning of LB-SOINN instead of clustering since our task is supervised. Therefore, we make the following adjustments to the original LB-SOINN.

\noindent1) The criteria to add a new node: Add a new node to the node set if one of the following condition is satisfied: 
a) The distance between the input and the winner is larger than the winner's threshold.
b) The distance between the input and the second winner is larger than the second winner's threshold.
c) The label of the input sample is not the same as the label of the winner.

\noindent2) Build connections between nodes: Connect the two nodes with an edge only if the winner and the second winner belong to the same class. 

\noindent3) We disable the removal of edges whose ages are greater than a predefined parameter.  We disable the deleting of nodes and the algorithm of updating the subclass labels of every node. The node label is the label of the instances assigned to it. Our adjusted algorithm guarantees that each node will only be assigned the samples from one class. 

LB-SOINN keeps track of the density of each node, which is defined as the mean accumulated points of a node. A node gets points when there is an input sample assigned to it. If the mean distance of the node from its neighbors is large, we give low points to the node. In contrast, if the mean distance of the node from its neighbors is small, we give high points to the node. Therefore, the density of the node reflects the number of nodes close to it and also the number of samples assigned to it. We take the density of the node as a measurement of the importance of the samples assigned to the node. After the LB-SOINN finishes training on the samples from the current task, we sort the samples according to the density of the node they are assigned to and the top $K$ samples are selected to write to the memory. We divide the memory equally for each of the previous tasks, so $K=M/t$, where $M$ is the total memory size and $t$ is the number of observed tasks, including the current task. The old memory consists of samples from the previous $t-1$ tasks and each task keeps $M/(t-1)$ samples in the old memory. For each of the $t-1$ tasks, the $M/(t-1)-M/t$ samples with the lowest node densities are deleted, resulting in $K$ empty slots in the memory, which is then rewritten by the selected $K$ samples in the current task.
\begin  {table*}[t!]
\centering
\small
\begin{tabular}{|l|c|c|c|c|c|c|}
  \hline
  Number of observed tasks & \multicolumn{2}{|c|}{t=5} &\multicolumn{2}{|c|}{t=10}& \multicolumn{2}{|c|}{t=15}\\
  \hline
  Avg F1 score (\%) &  Macro &  Micro &  Macro &  Micro &  Macro &  Micro \\
  \hline
  \hline
  Multitask &15.26 &67.07 &5.05 &37.20  &3.57  &38.61  \\
  \hline
  \hline
  Fine-tuning &6.02 &16.44 &4.35 &5.77  &3.96  &6.18  \\
  \hline
  Fine-tuning + BERT &6.02 &16.44 &4.06 &5.45  &3.03  &5.80  \\
  \hline
  Fine-tuning + RMR &11.15 &44.40 &2.56 &15.77  &3.51  &15.19  \\
  \hline

  EWC  &8.57 &20.42 &2.42 &6.81  &1.95  &7.27 \\
  \hline
  
  GEM  &\textbf{13.04} &30.95 &3.07 &12.51  &2.70  &15.07 \\
  \hline
  Ours &12.61 &\textbf{49.75} &\textbf{6.96} &\textbf{47.30}  &\textbf{5.13}  &\textbf{44.62} \\
  \hline
\end{tabular}
\caption{Experimental results. RMR: random memory replay. The best results are in bold. }
\label{tab:results}
\end{table*}

\begin{table*}[t!]
\centering
\small
\begin{tabular}{|l|c|c|c|c|c|c|}
  \hline
  Number of observed tasks & \multicolumn{2}{|c|}{t=5} &\multicolumn{2}{|c|}{t=10}& \multicolumn{2}{|c|}{t=15}\\
  \hline
  Avg F1 score (\%) &  Macro &  Micro &  Macro &  Micro &  Macro &  Micro \\
  \hline
  Full Model &12.61 &{49.75} &\textbf{6.96} &\textbf{47.30}  &{5.13}  &\textbf{44.62} \\
  \hline
  w/o $D_{KLmem}$ &\textbf{15.00} &\textbf{58.64} &4.21 &36.36  &3.72  &40.87  \\
  \hline
  w/o VRL &11.05 &35.03 &4.53 &13.69  &3.65  &11.28  \\
  \hline
  w/o LB-SOINN  &13.01 &50.99 &6.15 &44.42  &\textbf{5.59}  &30.91  \\
  \hline
\end{tabular}
\caption{Ablation study. w/o $D_{KLmem}$: $D_{KLmem}$ in the equation~\ref{eq:lossfinal} is removed. w/o VRL: VRL is replaced by the model used in the fine-tuning setting, i.e., fine-tuning + LB-SOINN memory replay. w/o LB-SOINN: LB-SOINN memory replay is replaced by random memory replay, i.e., VRL + RMR. The best results are in bold. }
\label{tab:ablation}
\end{table*}

\section{Experiments}
\label{sec:experiment}
\subsection{Experimental Settings}
For each task, we randomly sample 5000 tweets from the 80\% of the collected data for training, 10\% of the collected data for testing, and the rest 10\% for development. We allow the model to make more than one pass over the training samples in the current task or the current memory during training. We use average macro F1 score and average micro F1 score for evaluation. 
\begin{equation}
    \mbox{Average F1:} AvgF1(t)=\frac{1}{t}\sum_{i=1}^{t} F1_{t,i}
\end{equation}
where $F1_{t,i}$ is the F1 score, either macro F1 or micro F1, achieved by the model on the $i$th task after being trained on the $t$th task. The larger this metric, the better the model.
We compare our methods with the following methods:

\noindent{\bf Fine-tuning:}
The model contains two bidirectional LSTM encoders~\cite{hochreiter1997long, zhou2016attention, liu2016learning} to encode the tweet and the group separately. The score of the group is calculated as the cosine distance between the hidden state of the tweet encoder and that of the group encoder. This model is also the backbone model of all the methods described below, except Fine-tuning + BERT.
The model is directly fine-tuned on the stream of tasks, one after another, by the ranking loss function in~\ref{eq:lossmargin}.

\noindent{\bf Fine-tuning+BERT:} The training framework is the same as above, but each encoder is replaced by a pre-trained BERT model~\cite{devlin2019bert} followed by a linear layer.
The linear layers are fine-tuned during training.

\noindent{\bf Fine-tuning+RMR (Random Memory Replay):} We augment the fine-tuning method with an additional memory module. Same as in section~\ref{subsec:memory}, the memory is divided equally for each task, but instead of using LB-SOINN, the $K$ samples are randomly sampled from the current training data and then rewrite $K$ random slots in the old memory.

\noindent{\bf EWC:} EWC is a regularization-based method, adding a penalty term $\sum_i\frac{\lambda}{2}F_i(\theta_i-\theta^*_{i})^2$ to the  ranking loss function~\ref{eq:lossmargin}.  $F_i$ is the diagonal of the Fisher information matrix $F$, $\theta$ is the model parameter, and $i$ labels each parameter. $\theta^*$ is the model parameter when the model finishes training on the previous task. $\lambda$ is set to 2e6 in our experiments.

\noindent{\bf GEM:} We use the episodic memory in the original paper: the memory is populated with $m$ random samples from each task. $m$ is a predefined size of the episodic memory. We set $m=100$ in our experiments, so each task can add 100 tweets to the memory. By the end of the 15 tasks, the total memory of GEM contains 1500 tweets. 

\noindent{\bf Multitask Learning:}  The tasks are trained simultaneously. We mix the training data from multiple tasks to train the model. This setting does not follow the lifelong learning setting where the tasks are trained sequentially. We add this setting in our experiments to show the potential room for improvement concerning each lifelong learning method.

We do not compare our method with Support Vector Machine~\cite{suykens1999least} or Logistic Regression, because they require the number of classes to be fixed and to be known in advance, which is unrealistic in our tasks. We also do not compare our method with~\citet{qian2018hierarchical} since the latter also has this requirement, as mentioned in section~\ref{sec:desc}. Adapting their method for the lifelong learning setting requires modifying both the model architecture and the training algorithm, which is beyond the scope of this paper.

In all our experiments, we use 1-layer bi-LSTM as encoders except the fine-tuning + BERT setting and we use cosine distance to measure similarity. The input of the group encoder is the concatenation of the group name and its hate ideology. 
We use 1-layer bidirectional GRU~\cite{cho2014learning} as the decoder in VRL. The hidden size of the encoders and the decoders is 64. The latent variable size in VRL is 128. We use 300-dimensional randomly initialized word embeddings. All the neural networks are optimized by Adam optimizer with the learning rate 1e-4. The batch size is 64. The loss margin $m=0.5$. The maximum number of training epochs for each task is set to 20. For LB-SOINN, $\lambda\!\!=\!\!1000$, $\eta\!\!=\!\!1.04$.
The memory size is limited to 1000 tweets for all the methods using a memory module except GEM. We do not set episodic memory size for each task as GEM because for lifelong hate speech classification, the number of tasks keeps increasing in the real world, and assuming unlimited total memory is unrealistic. 
\subsection{Experimental Results}
The experimental results are shown in Table~\ref{tab:results}. We report the performance of each method after the model finishes training on the first 5 tasks, first 10 tasks, and all the 15 tasks. The average macro-F1 score is much lower than the average micro-F1 score due to the imbalanced data of each task. 
The large performance gap between the multitask training and fine-tuning shows that there exists severe catastrophic forgetting and that the low average F1 scores in the fine-tuning setting are not due to the model capacity. Replacing the bi-LSTM encoder with the pre-trained BERT encoder does not improve the performance.
This reconfirms that the low scores result from catastrophic forgetting, not model capacity. Actually fine-tuning and fine-tuning with BERT achieves the same average F1 scores at $t=5$ because both models completely forget the previous tasks after converging on the fifth task, so both models achieve the same F1 scores on the testing data of the fifth task while achieving 0 scores on the previous four tasks. Due to the large model capacity of BERT, fine-tuning with BERT tends to overfit on the training data more seriously, leading to slight performance decline at $t=10$ and $t=15$ compared to using bi-LSTM encoders. Since model capacity is not the key factor to solve catastrophic forgetting, we simply use bi-LSTM as encoders in our model instead of BERT, considering the computational cost.

Adding RMR to the fine-tuning setting achieves significant performance improvement, even better than EWC or GEM. This is related to the characteristic of our tasks mentioned at the end of section~\ref{sec:desc}. EWC remembers previous tasks by slowing down the update of the model parameters important to them, which is more suitable for the sequence of tasks that are similar to each other. However, significant changes in vocabulary, topic, or input data distribution are very common in our sequence of tasks, making memory replay more efficient than EWC. The performance of GEM during the second half of the training is close to that of fine-tuning with RMR, but there exists a gap in the first half. The reason is that GEM sets an episodic memory for each task, of which the size is 100 in our experiments, so before the 10th task in the sequence, the size of the total memory available for GEM is less than that of the memory module used in the fine-tuning with RMR setting.   

Although RMR improves the performance, the average F1 scores still drop quickly when the number of tasks increases. In the late stage of sequential training, each task can only keep dozens of samples in the memory and the model is not able to generalize well based on the memory. Our method solves this problem by combining VRL and LB-SOINN memory replay. The performance of our model is better and more stable than the other methods when the number of tasks increases. Our method achieves higher scores than multitask training in the last four columns of Table~\ref{tab:ablation} because learning on one task is easier than learning on a mix of tasks simultaneously. Every model in our sequential training experiments can easily achieve high F1 scores on the current task, making a large contribution to the average F1 scores. However, when doing multitask training, the model loses this benefit. 

To investigate the effect of our method, we conduct the ablation study as shown in Table~\ref{tab:ablation}. 
Removing $D_{KLmem}$ from the final loss function in equation~\ref{eq:lossfinal} does not lower the performance when the number of observed tasks is small ($t\!\!=\!\!5$) because each task can store a few hundreds of samples in the memory at the early stage of sequential training, which is sufficient for the model to learn the previous tasks. However, when the number of tasks increases, $D_{KLmem}$ shows its effect on alleviating catastrophic forgetting. 

Fine-tuning+LB-SOINN (Table~\ref{tab:ablation}) does not perform as well as fine-tuning+RMR (Table~\ref{tab:results}), while VRL+LB-SOINN (i.e., full model) performs better than VRL+RMR (Table~\ref{tab:ablation}). The reason lies in the input for LB-SOINN.
Compared to the hidden representations spread evenly in the hidden space, the hidden representations which are well-organized in different group clusters make it easier for LB-SOINN to learn a reasonable topology structure of the training samples. VRL achieves this by explicitly pushing the hidden representation of tweets to follow a learned multivariate Gaussian distribution unique to each group. On the other hand, directly using the hidden state of the tweet encoder does not exhibit such kind of characteristics. VRL not only distills task knowledge but also provides an appropriate input for LB-SOINN, as stated in section~\ref{sec:approach}.
\begin{figure}[t]
\centering
\includegraphics[width=0.4\textwidth]{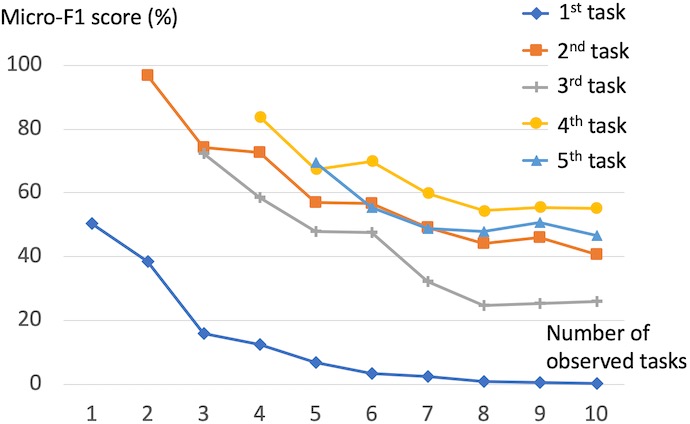}
\caption{The testing results of the first 5 tasks in the sequence when our model is trained on the first 10 tasks.}
\label{fig:error}
\end{figure}
\subsection{Error Analysis}
Although our model achieves significant improvement over the baseline methods, we observe that our method does not perform well on the first task. As shown in Figure~\ref{fig:error}, there exists a large gap between the performance on the first task and the other tasks, and the micro-F1 score on the first task quickly drops to almost 0 when the number of observed tasks increases. We find the same results after we change the order of tasks in the sequence, so this is not the result of the task difficulty but is the result of our method. 
We find this problem is due to the reconstruction loss, which is the first part in equation~\ref{equ:lossvae}.
The model observes a very limited number of tweets when training on the first task, making it difficult to learn the language model and reconstruct the tweet. As a result, the tweet representation learned on the first task may not contain the information we require, resulting in a large performance gap. When the number of observed tasks increases, this problem goes away quickly. We anticipate pre-training the VAE in our model (the left branch in Figure~\ref{fig:model}) on a large Twitter corpus can alleviate this problem at the beginning of training.

\section{Conclusion}
In this paper, we introduce the lifelong hate speech classification task and propose to use the VRL and LB-SOINN memory module to alleviate catastrophic forgetting. 
Our proposed method has the potential to benefit other lifelong learning tasks where the similarity between the contiguous tasks can be low. We make our implementation freely available to facilitate more application and investigation of our method in the future\footnote{https://www.aclweb.org/anthology/2021.naacl-main.183/}.
\bibliographystyle{acl_natbib}
\bibliography{anthology,custom}

\appendix
\end{document}